\documentclass[journal]{IEEEtran}

\ifCLASSINFOpdf
\else
   \usepackage[dvips]{graphicx}
\fi
\usepackage{balance}
\usepackage{url}
\usepackage{graphicx}
\usepackage{cite}
\usepackage{amsmath,amssymb,amsfonts}
\usepackage{algorithmic}
\usepackage{textcomp}
\usepackage{subfigure}
\usepackage{caption}
\usepackage{xcolor}
\usepackage{booktabs}
\usepackage{arydshln}
\usepackage{bbding}
\usepackage{multirow}
\usepackage{diagbox}
\usepackage{ntheorem}
\usepackage{makecell}
\usepackage{colortbl}
\usepackage{pifont}%
\usepackage{tablefootnote}

\usepackage[pagebackref=true,breaklinks=true,colorlinks,bookmarks=false]{hyperref}
\usepackage[ruled,vlined]{algorithm2e}
\usepackage{multirow}
\usepackage{diagbox}
\usepackage{textcomp}
\usepackage{subfigure}
\usepackage{caption}
\usepackage{xcolor}
\usepackage{booktabs}
\usepackage{arydshln}
\usepackage{bbding}
\usepackage{booktabs}
\usepackage{array}
\usepackage{epsfig}
\usepackage[font={bf},textfont=md]{caption}
\usepackage{float}
\usepackage{stfloats}
\usepackage[misc]{ifsym}

\usepackage[font={bf},textfont=md]{caption}

\usepackage[switch]{lineno} 
\usepackage{siunitx}
\newcommand{\figref}[1]{Fig.~\ref{#1}}
\newcommand{\equref}[1]{Eq.~\eqref{#1}}

\newcommand{\tabref}[1]{Table.~\ref{#1}}

\makeatletter
\DeclareRobustCommand\onedot{\futurelet\@let@token\@onedot}
\def\@onedot{\ifx\@let@token.\else.\null\fi\xspace}

\def\eg{\emph{e.g}\onedot} 
\def\ie{\emph{i.e}\onedot}

\makeatother

\definecolor{seagreen}{RGB}{84,255,159}
\definecolor{SpringGreen}{RGB}{0,139,69}

\newcommand{\PreserveBackslash}[1]{\let\temp=\\#1\let\\=\temp}
\newcolumntype{C}[1]{>{\PreserveBackslash\centering}p{#1}}
\newcolumntype{R}[1]{>{\PreserveBackslash\raggedleft}p{#1}}
\newcolumntype{L}[1]{>{\PreserveBackslash\raggedright}p{#1}}

\begin{document}

\title{Searching Dense Point Correspondences via Permutation Matrix Learning}

\author{Zhiyuan Zhang$^1$, Jiadai Sun$^1$, Yuchao Dai$^1$\textsuperscript{\Letter}, Bin Fan$^1$, and Qi Liu$^1$

\thanks{This work was supported in part by the National Key Research and Development Program of China (2018AAA0102803) and National Natural Science Foundation of China (61871325, 61901387), and Innovation Foundation for Doctor Dissertation of Northwestern Polytechnical University (CX2022046).
}
\thanks{$^1$School of Electronics and Information, Northwestern Polytechnical University, Xi'an, 710129, China. Yuchao Dai is the corresponding author, Email: daiyuchao@nwpu.edu.cn.}
}

\markboth{Journal of \LaTeX\ Class Files, Vol. 14, No. 8, August 2015}
{Shell \MakeLowercase{\textit{et al.}}: Bare Demo of IEEEtran.cls for IEEE Journals}

\maketitle

\begin{abstract}

Although 3D point cloud data has received widespread attentions as a general form of 3D signal expression, applying point clouds to the task of dense correspondence estimation between 3D shapes has not been investigated widely.
Furthermore, even in the few existing 3D point cloud-based methods, an important and widely acknowledged principle, \ie one-to-one matching, is usually ignored. 
In response, this paper presents a novel end-to-end learning-based method to estimate the dense correspondence of 3D point clouds, in which the problem of point matching is formulated as a zero-one assignment problem to achieve a permutation matching matrix to implement the one-to-one principle fundamentally.
Note that the classical solutions of this assignment problem are always non-differentiable, which is fatal for deep learning frameworks. Thus we design a special matching module, which solves a doubly stochastic matrix at first and then projects this obtained approximate solution to the desired permutation matrix. Moreover, to guarantee end-to-end learning and the accuracy of the calculated loss, we calculate the loss from the learned permutation matrix but propagate the gradient to the doubly stochastic matrix directly which bypasses the permutation matrix during the backward propagation.
Our method can be applied to both non-rigid and rigid 3D point cloud data and extensive experiments show that our method achieves state-of-the-art performance for dense correspondence learning. 

\end{abstract}


\IEEEpeerreviewmaketitle

\section{Introduction}\label{sec:introduction}
As a new form of 3D signal expression, point cloud has received widespread attention due to the talent for representing 3D shapes efficiently and being obtained directly and easily by scanning. And it has been adopted in many applications, such as segmentation \cite{GANet_Deng_spl_21,segmentation_zhang_spl_20},
3D reconstruction \cite{deschaud_imls_icra_2018,zhang_loam_rss_2014}, compression \cite{compression_evaristo_spl_21,compression_gu_spl_21,compression_peixoto_spl_21}, autonomous driving \cite{wan_robustlocalization_icra_2018}.
Particularly, although dense correspondence estimation is a fundamental task in computer vision, which has been investigated widely on other signal formats (\eg 2D image \cite{matching_li_spl_21,rsdpsnet_spl_21}), it remains a challenging problem for point cloud data due to the inherent characteristics of the point cloud (\eg irregularity, orderless), the serious deformation of object shape, and so on. 
Note that most existing methods of 3D shape correspondence estimation pay more attention to the 3D mesh input \cite{oliver_summary_cgf_2011}. And although many 3D mesh-based methods have been encouraged \cite{litany_deepFM_iccv_2017,halimi_unfmnet_cvpr_19,donati_deepgfm_cvpr_20,boscaini_lsc_nips_16,monti_gdl_cvpr_17}, these methods cannot be extended to 3D point cloud input directly since there is no topology information existing in 3D point cloud comparing with the 3D mesh.

To solve the correspondence searching problem for the 3D point clouds directly, scene flow estimation methods \eg FlowNet3D \cite{liu_flownet3d_cvpr_19}, are usually borrowed by selecting the nearest point as the corresponding point. However, an important and widely acknowledged principle, \ie one-to-one matching, is ignored. This will result in a large number of one-to-many matches and unsatisfactory performance.
CorrNet3D \cite{zeng_corrnet3d_cvpr_2021} is a representative method advocating solving a permutation matching matrix to ensure the one-to-one matching. However, this intention is not realized because the learned matching matrix is still a soft probability matrix and the final corresponding point is determined by selecting the point with the largest matching probability where the one-to-many matching is still not excluded. 
Besides, as a special task, learning point correspondence between rigid point clouds has been broadly investigated as a crucial intermediate step in the point cloud registration field \cite{zhang_review_vrih_2020}. 
However, most of them \cite{choy_dgr_cvpr_2020,D3Feat_Bai_2020_CVPR,perfect_Gojcic_cvpr19,wang_prnet_nips_2019} advocate building some reliable correspondences, which is reasonable for registration but do not make sense for dense correspondence estimation task. Some virtual point-based methods \cite{wang_dcp_iccv_2019,yew_rpmnet_cvpr_2020} advocate estimating dense correspondence, however, the learned correspondence is not accurate since the constructed virtual points degenerate seriously.
In summary, as far as we know, the one-to-one matching principle has not been implemented well in the existing 3D point cloud-based dense correspondence learning methods.

In this paper, we focus on the dense correspondence estimation problem of 3D shapes in the form of point clouds and achieve the real one-to-one matching learning in an end-to-end manner for the first time. 
To this end, we formulate the point matching as a zero-one assignment problem to solve a special binary matrix, permutation matrix (PM) \footnote{The formal definitions of PM, DSM are given in the following section.} to implement one-to-one matching principle fundamentally. Although the binary matrix has been investigated widely in \cite{li_wire_ijcv_2020,li_wmre_tmm_2015}, the desired PM further constrains the sum of each row and column. Note that the classical solutions of the assignment problem are always not differentiable \cite{kuhn1955hungarian}, which is fatal for deep learning frameworks. In response, a special matching module is designed.
This module firstly solves a doubly stochastic matrix (DSM) as the approximate solution and then projects the obtained DSM to the desired PM. Moreover, to guarantee the end-to-end learning and the accuracy of the calculated loss, we advocate calculating the loss from the learned PM but propagating the gradient to the DSM directly during the backward propagation.
Our method can be employed to both non-rigid and rigid point cloud data. 
Extensive evaluations on several benchmark datasets validate that our method achieves state-of-the-art performance for dense correspondence learning. 

\begin{figure}[!t]
	\centerline{\includegraphics[width=0.5\textwidth]{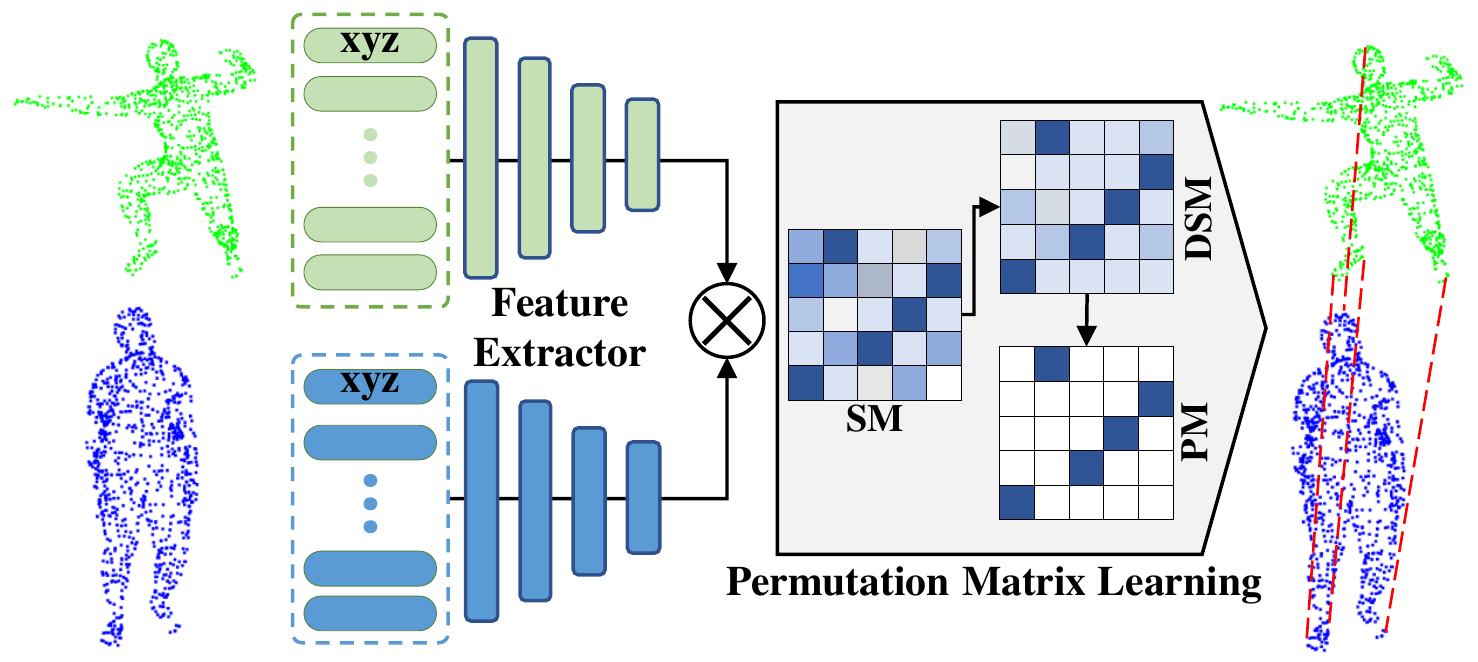}}
	\vspace{-0.25cm}
	\caption{The outline of the proposed dense correspondence learning framework. For the input 3D point clouds, the point features are extracted at first, and then the one-to-one matching is achieved by learning a permutation matching matrix. %
	}
	\label{fig:network}
	\vspace{-0.6cm}
\end{figure}

Our contributions can be summarized as follows:
1). We propose to learn the permutation matching matrix in an end-to-end learning to implement one-to-one principle fundamentally. 2). Our method can handle both non-rigid and rigid 3D point clouds directly for dense correspondence estimation. 3). Extensive evaluations validate the superiority of our method, which creates a new state-of-the-art performance.

\section{Method}\label{sec:Proposed_Approach}
In this section, our dense correspondence learning approach is introduced in detail. The outline is illustrated in \figref{fig:network}. 

\vspace{-0.3cm}
\subsection{Preliminaries}\label{sec:preliminaries}

Given the \textit{source} point cloud $\mathbf{X}= [\mathbf{x}_{i}] \in \mathbb{R}^{3\times N}$ and the \textit{target} point cloud $\mathbf{Y}= [\mathbf{y}_{j}] \in \mathbb{R}^{3\times N}$, which represent the same deformable 3D shape, we aim at estimating the dense point-to-point correspondence in this paper. Specifically, the matching matrix $\mathbf{M} = [m_{ij}] \in \{0, 1\}^{{N} \times {N}}$ is searched to indicate the assignment between the points in $\mathbf{X}$ and $\mathbf{Y}$, where $m_{ij}=1$ if point $\mathbf{x}_i$ in $\mathbf{X}$ and point $\mathbf{y}_j$ in $\mathbf{Y}$ are matched, and $m_{ij}=0$ otherwise. 
The desired $\mathbf{M}$ is equivalent to PM, \ie $\sum_{i=1}^N m_{ij} = 1, \forall j$, $\sum_{j=1}^N m_{ij} = 1, \forall i$ and $m_{ij} \in \{0, 1\}$, by which the one-to-one matching principle can be formulated.

\vspace{-0.3cm}
\subsection{Feature Extractor}\label{sec:Feature_extractor}
For accurate matching, distinguished descriptors are crucial. Currently, deep features have been widely used. In our method, we apply ``DGCNN + Transformer'' as the feature extractor. Specifically, a DGCNN \cite{wang_dgcnn_tog_2019} is firstly employed to compute the initial point features, which incorporates the neighbor information in both 3D Euclidean space and feature space. Besides, inspired by the recent success of the attention mechanism, we use Transformer \cite{vaswani_attention_nips_2017} to learn cross-contextual information for more descriptive point features. Formally, we denote the DGCNN feature as $\mathbf{F}_\mathbf{X}=[\mathbf{F}_{\mathbf{x}_i}]\in \mathbb{R}^{N\times d}$, $d$ is the pre-defined feature dimension. Analogously, $\mathbf{F}_\mathbf{Y}=[\mathbf{F}_{\mathbf{y}_j}]\in \mathbb{R}^{N\times d}$. Then, the Transformer is applied, \ie $\Phi_\mathbf{X} = \mathbf{F}_\mathbf{X} + \eta(\mathbf{F}_\mathbf{X}, \mathbf{F}_\mathbf{Y})$, $\Phi_\mathbf{Y} = \mathbf{F}_\mathbf{Y} + \eta(\mathbf{F}_\mathbf{Y}, \mathbf{F}_\mathbf{X})$,
where $\eta:\mathbb{R}^{N\times d} \times \mathbb{R}^{N\times d} \to \mathbb{R}^{N \times d}$ is the Transformer function. The final point features of all \textit{source} points and \textit{target} points are represented as $\Phi_\mathbf{X}\!=\![\Phi_{\mathbf{x}_i}] \in \mathbb{R}^{N\times d}$ and $\Phi_\mathbf{Y}\!=\![\Phi_{\mathbf{y}_j}]\in \mathbb{R}^{N\times d}$.

\vspace{-0.3cm}
\subsection{Permutation Matrix Learning}
\label{sec::mathcing}
As mentioned above, we devote to learning a permutation matrix to implement one-to-one principle fundamentally. To this end, we design a dedicated matching module as follows to guarantee the end-to-end learning.

\vspace{0.5mm}
\noindent\textbf{Initial similarity matrix solving.}
In our matching module, we firstly compute an initial similarity matrix (SM). Based on the obtained deep features ${\Phi_\mathbf{X}}$ and ${\Phi_\mathbf{Y}}$, the SM $\mathbf{S}$ is calculated by using the scale dot product attention metric \cite{vaswani_attention_nips_2017}, \ie
$\mathbf{S} = \Phi_\mathbf{X} \Phi_\mathbf{Y}^\text{T} / \sqrt{d}$,
where the entry $s_{ij}$ represents the similarity between point $\mathbf{x}_i$ in $\mathbf{X}$ and point $\mathbf{y}_j$ in $\mathbf{Y}$. 

\vspace{0.5mm}
\noindent\textbf{Permutation matrix Learning.}
Note that the SM $\mathbf{S}$ is in continuous space while the desired PM $\mathbf{M}$ is in discrete space with the entry of either 0 or 1, solving PM can be defined as a zero-one assignment problem whose existing solutions are always non-differentiable. Therefore, how to optimize $\mathbf{S}$ to $\mathbf{M}$ in a deep learning pipeline is the main challenge.
To solve this problem, we design an ingenious matching method as presented in Alg. \ref{alg:algorithm1}. And a DSM $\mathbf{P}$ is solved as an intermediate variable between $\mathbf{S}$ and $\mathbf{M}$ as shown in \figref{fig:network}, where $\mathbf{P}=[p_{ij}]\in {[0,1]}^{N\times N}$, $\sum_{i=1}^N p_{ij} = 1, \forall j$, $\sum_{j=1}^N p_{ij} = 1, \forall i$. Firstly, this two-stage method applies the Gumbel-Sinkhorn algorithm \cite{gonzalo_gumbelsinkhorn_iclr_2018} to solve $\mathbf{P}$ as an approximate solution of $\mathbf{M}$. Compared to the original Sinkhorn method, the Gumbel-Sinkhorn is more robust by adding a standard i.i.d. Gumbel noise matrix $\boldsymbol\epsilon$ to the input $\mathbf{S}$. Then, $\mathbf{P}$ is projected to the final permutation matching matrix $\mathbf{M}$ by formulating a zero-one assignment problem, in which objective function is defined by considering the DSM $\mathbf {P}$ as the profit matrix, \ie 
\begin{equation} \vspace{-0.2cm}
\mathbf{M}^{*} = \arg \mathop {\max }\limits_{\mathbf{M} \in \mathcal{M}_{N}}  {<\mathbf{M},\mathbf{P}>}_{F},
\label{obj2}
\end{equation}
where $\mathcal{M}_{N}$ denotes the set of permutation matrices and ${<\mathbf{M},\mathbf{P}>}_{F}=trace(\mathbf{M}^{\mathbf{T}} \mathbf{P})$ denotes the (Frobenius) inner product of matrices. There are many classical solutions to this assignment problem. In our implementation, the representative Hungarian algorithm \cite{kuhn1955hungarian} is applied.

\begin{algorithm}[!h]
\algsetup{linenosize=\small} \small
	\KwIn{Initial similarity matrix $\mathbf{S}$, Gumbel noise matrix $\boldsymbol\epsilon$, Hyper-parameter $\mu$, iteration number $Iter$.}
	\KwOut{Optimal Permutation Matrix  $\mathbf{M}^{*}$.}  
	\BlankLine
	$\mathbf{S}=(({\mathbf{S} +\boldsymbol\epsilon})/ \mu)$;  \textcolor{gray}{// add Gumbel noise to similarity matrix }
	
	$\mathbf{S}=\text{exp}(\mathbf{S})$;
	
	\While{$it \leqslant Iter$}{
		${\mathbf{S}={\mathcal{T}_r}(\mathbf{S})}$;  \textcolor{gray}{// ${\mathcal{T}_r}$ indicates row-wise normalization}  
		
		${\mathbf{S}={\mathcal{T}_c}(\mathbf{S})}$;  \textcolor{gray}{// ${\mathcal{T}_c}$ indicates column-wise normalization}  
		
	} 
	
	$\mathbf{P}=\mathbf{S}$; \textcolor{gray}{// $\mathbf{P}$ is the DSM returned by final $\mathbf{S}$}.
	
	$\mathbf{M}^{*}={h}(\mathbf{P})$. \textcolor{gray}{// $h(\cdot)$ indicates the Hungarian algorithm}
 \caption{Permutation Matrix Learning.}
 \label{alg:algorithm1}
\end{algorithm}

\subsection{Loss Function}\label{subsec:loss_fun}
Here, we supervise the learned matching matrix directly. Specifically, the loss function is defined as
\begin{equation}
\mathcal{L} =  - \frac{1}{N}{\sum_{i=1}^N \sum_{j=1}^N \left( {{m}_{ij}^\text{pred} {m}_{ij}^\text{gt}} \right)},
\label{eq:loss}
\end{equation}
where the superscript ``$\text{pred}$'' and ``$\text{gt}$'' indicate the prediction and the ground truth, respectively. %

\vspace{-0.3cm}
\subsection{Remarks} \label{subsec:remark}
\begin{figure}[!t]
	\centerline{\includegraphics[width=0.35\textwidth]{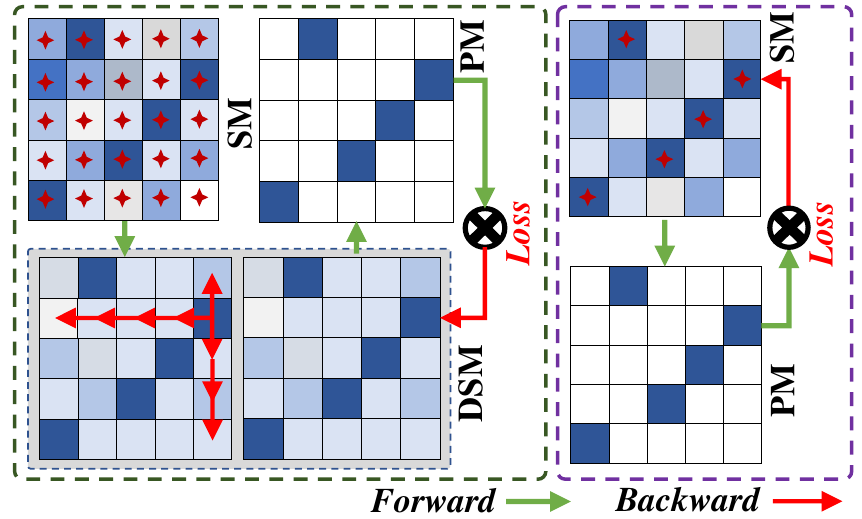}}
	\vspace{-0.2cm}
	\caption{The illustration of forward and backward propagation in our two-stage matching module (Left) and one-stage matching module (Right). Our method not only guarantees the end-to-end learning but also avoids the local supervision risk.
	}
	\label{fig:com}
	\vspace{-0.6cm}
\end{figure}

Here, we stress the ingenuity of the designed PM learning module in the end-to-end learning pipeline.
As mentioned above, this zero-one assignment problem is fatal for the deep learning pipeline due to the non-differentiability. 
As a response, we propose a ingenious operation as shown in \figref{fig:com} Left. Specifically, during the forward propagation, the loss is calculated based on the learned PM $\mathbf{M}$. 
However, during the backward propagation, the gradient is propagated to the learned DSM $\mathbf{P}$ directly bypassing the PM solving procedure. This operation guarantees the accuracy of the calculated loss and the end-to-end training synchronously. %

Then, a natural question is that, \textit{since the assignment procedure is bypassed in backward propagation, why not use the similarity matrix as the profit matrix to solve the permutation matrix but use DSM as the profit matrix}. Although this one-stage strategy is more direct as shown in \figref{fig:com} Right, it exists an obvious local supervision risk. Specifically, if the gradient is propagated to the SM directly, the correlation of entries, which should have been considered in the assignment algorithm, is ignored in backward propagation, \ie the loss calculated from $m_{ij}$ can only supervise $s_{ij}$. Note that, from \equref{eq:loss}, when $m_{ij}^\text{gt}=1$, $m_{ij}^\text{pred}$ is enforced to 1. When $m_{ij}^\text{gt}=0$, $m_{ij}^\text{pred}$ is trained without supervision. Thus many entries (${N}({N} - 1)$) will be trained without supervision in this one-stage method, and the training will not converge. %
Our solution avoids this local supervision risk effectively by propagating the gradient to the intermediate DSM, where the correlation of all entries is considered again in Gumbel-Sinkhorn. Through our method, the similarity of the correct correspondences will be boosted, meanwhile the similarity of the wrong correspondences will be effectively suppressed during the backward propagation.

\section{Experimental Results}
\label{sec:expriments}
We conduct extensive experiments on both non-rigid and rigid point cloud. In our application, in the feature extractor, a 4-layer graph neural network is applied where the parameter of K-nearest neighbor is set to 24, the embedding dimension $d=512$, iteration number $Iter=5$, hyper-parameter $\mu=0.5$. As for the training, we at first train the feature extractor with an incomplete pipeline, \ie solving DSM only without assignment for 100 epochs. Then, we insert the Hungarian algorithm, and retrain the entire network for 100 epochs. The initial learning rate in the Adam optimizer is set to 0.001. 

\noindent\textbf{Datasets.} 
For non-rigid shape correspondence, following \cite{zeng_corrnet3d_cvpr_2021}, we adopted Surreal \cite{groueix_3dcoded_eccv_18} and SHREC \cite{donati_deepgfm_cvpr_20} as the training and test sets respectively. 
Besides, the re-meshed versions of FAUST \cite{bogo_faust_cvpr_14} and SCAPE \cite{anguelov_scape_tog_2005} are also used following \cite{arnold_detection_tits_19}.
For all mesh data, our method randomly picked vertices as the point cloud.
For rigid shape correspondence, ModelNet40 \cite{wu_modelnet40_cvpr_2015} is used following the registration methods \cite{wang_dcp_iccv_2019,yew_rpmnet_cvpr_2020}.

\noindent\textbf{Metrics.}  
We use the matching precision (\%) metric here to evaluate the dense correspondence estimation performance. Theoretically, it presents the correct correspondence percentage by $\frac{1}{N} \|\mathbf{M}^\text{pred} \odot \mathbf{M}^\text{gt}\|_1$,
where $\odot$ is the Hadamard product, $\mathbf{M}^\text{pred}$, $\mathbf{M}^\text{gt}$ represent the predicted and ground truth matching matrices, respectively. In actual application, this metric is often relaxed by confirming the correspondence as a correct result when the distance between the predicted corresponding point and the real corresponding point is less than a threshold. Here, we propose a self-adaptive threshold as
${\tau _i} = \frac{1}{K} {\sum_{j = 1}^K {{{l}_{\mathbf{x}_{i},\mathbf{x}_j}}}}$, $\mathbf{x}_{j} \in \text{KNN} \left( \mathbf{x}_{i} \right)$,
where ${l_{\mathbf{x}_{i},\mathbf{x}_{j}}}$ is the Euclidean distance between the points $\mathbf{x}_{i}$ and $\mathbf{x}_{j}$, $\text{KNN}(\cdot)$ represents the K-nearest neighbor. In other words, for $i$-th point $\mathbf{x}_{i}$, $\tau_i$ is computed as the mean distance of $K$ nearest neighbor points.
Meanwhile, for a fair comparison, the metric of per-point-average geodesic distance \cite{donati_deepgfm_cvpr_20} is also used for the methods with 3D mesh input.

\noindent\textbf{Comparison methods:} 
For non-rigid shape correspondence estimation, we first compare with point cloud-based methods including \underline{F}lowNet\underline{3D} \cite{liu_flownet3d_cvpr_19}, \underline{CorrNet3D} \cite{zeng_corrnet3d_cvpr_2021}, in \underline{U}nsupervised and \underline{S}upervised training, notated as U-F3D, S-F3D, U-CorrNet3D and S-CorrNet3D respectively. 
Then, 3D mesh-based methods are also compared, including traditional methods ZoomOut \cite{melzi_zoomout_tog_19}, BCICP\cite{ren_bcicp_tog_18}, and learning-based methods 3D-CODED \cite{groueix_3dcoded_eccv_18}, FMNet \cite{litany_deepFM_iccv_2017}, U-FMNet\cite{halimi_unfmnet_cvpr_19} with and without post-processing (PMF \cite{pmf_vestner_cvpr_17}), SURFMNet \cite{roufosse_surfmnet_iccv_19} with and without ICP, DeepGFM \cite{donati_deepgfm_cvpr_20} with and without ZoomOut (ZO).
For rigid input, ICP \cite{besl_icp_pami_1992}, DCP \cite{wang_dcp_iccv_2019}, PRNet \cite{wang_prnet_nips_2019} are evaluated. 
For PRNet, two versions have been evaluated here. For the original version, which is denoted as PRNet, half input points are selected as key points. 
For the modified version, which is denoted as PRNet*, all points are considered as key points for dense correspondence.

\begin{figure}
\begin{minipage}{.45\linewidth}
    \centering
    \includegraphics[width=4.1cm,height=3.8cm]{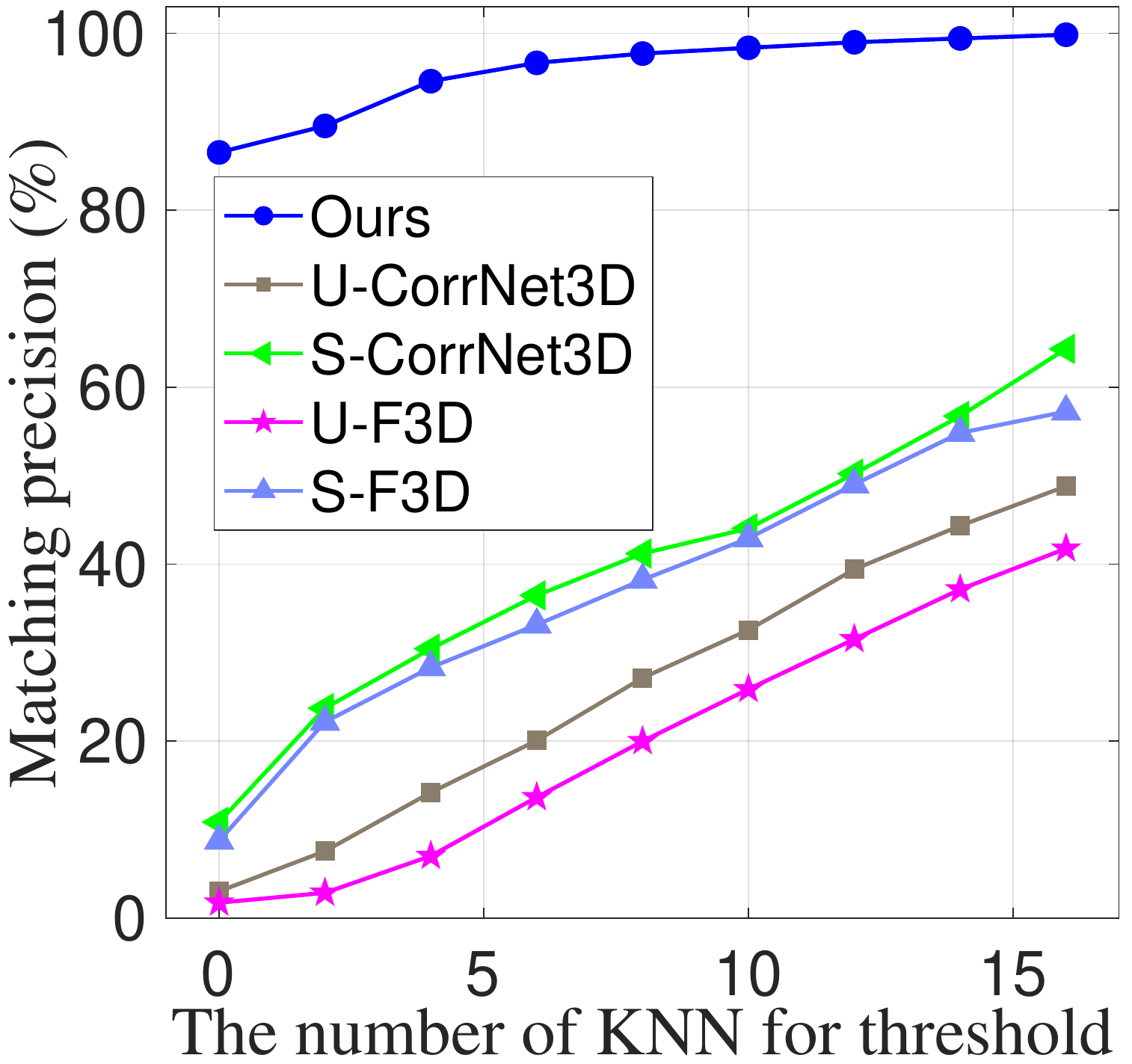}
	\caption{Comparison of the matching precision for non-rigid 3D point clouds on Surreal\textbackslash SHREC. And $K=0$ means the threshold is 0.}
	\label{fig:human_precision}
\end{minipage}
\begin{minipage}{.52\linewidth}
	\begin{center}
	    \resizebox{120pt}{!}{
			\begin{tabular}{lccccc}
				\toprule
				\textbf{Method } & \textbf{F}\textbackslash \textbf{F} & \textbf{S}\textbackslash \textbf{S} & \textbf{F}\textbackslash \textbf{S} & \textbf{S}\textbackslash \textbf{F}   \\  
				\midrule
				\textbf{BCICP}          & 15. &  16.  &  $\ast$ &  $\ast$ \\
				\textbf{ZoomOut}        & 6.1 &  7.5  &  $\ast$ &  $\ast$ \\ 
				\midrule
				\textbf{SURFMNet}       & 15. &  12.  & 32.   &  32.  \\
				\textbf{SURFMNet}+\textbf{icp} & 7.4 & 6.1 & 19. & 23. \\
				\textbf{U-FMNet}    & 10. &  16.  & 29.   &  22.   \\
				\textbf{U-FMNet}+\textbf{pmf} &5.7 & 10. & 12. & 9.3 \\
				\textbf{FMNet}          & 11. & 17. & 30. & 33. \\
				\textbf{FMNet}+\textbf{pmf} &5.9 & 6.3 & 11. & 14.\\
				\textbf{3D-CODED}       & 2.5 &31. & 31. & 33. \\
				\textbf{DeepGFM}        & 3.1 & 4.4 & 11. & 6.0\\
				\textbf{DeepGFM}+\textbf{zo} & {1.9} & \textbf{3.0} & \textbf{9.2} & \textbf{4.3}\\
				\midrule
				\textbf{Ours}           &\textbf{1.7} & 4.6 & {9.3} & 5.2 \\
				\bottomrule
			\end{tabular}}
			\captionof{table}{Comparison of per-point-average geodesic distance results ($\times 100$). Four dataset settings are evaluated, which are indicated by \textbf{train set}\textbackslash \textbf{test set}.} 
		\label{tab:fs}
	\end{center}
\end{minipage}
\vspace{-0.5cm}
\end{figure}

\vspace{-0.2cm}
\subsection{Evaluation on Non-rigid Data}
\begin{figure}[!t]
	\centerline{\includegraphics[width=0.5\textwidth]{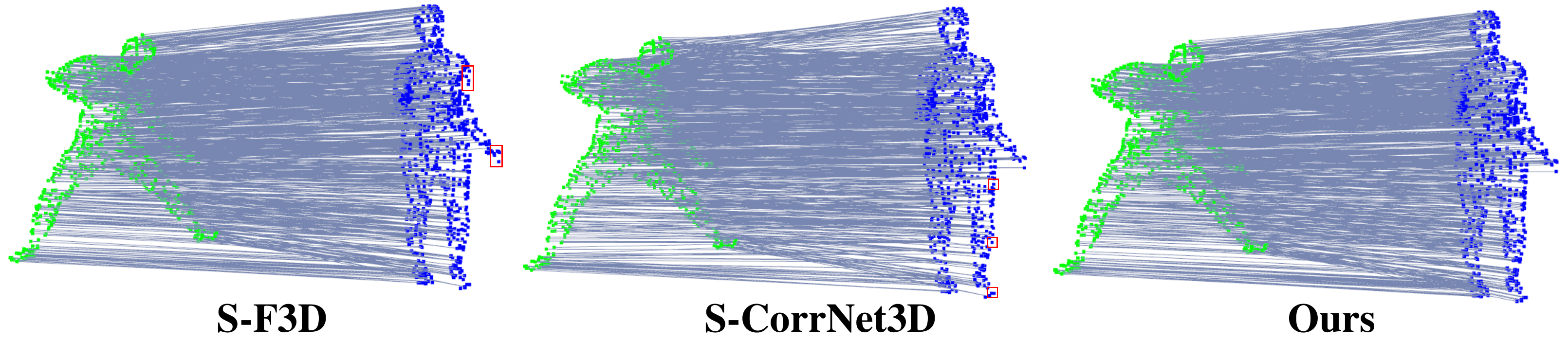}}
	\vspace{-0.2cm}
	\caption{Matching results on SHREC, where green and blue represent the \textit{source} and \textit{target} respectively. The points which are not assigned corresponding points (indicated by the red box) show that one-to-many matching exists in baselines.}
	\label{fig:human_point_cloud}
	\vspace{-0.5cm}
\end{figure}

\figref{fig:human_precision} shows the quantitative comparison with 3D point clouds input on Surreal\textbackslash SHREC. Our method always produces the best performance under different thresholds, which outperforms all baselines by a large margin. Meanwhile, S-CorrNet3D achieves second-top results, which is better than S-F3D. We speculate that the large deformation of input shapes is difficult for scene flow estimation. Besides, we present a qualitative comparison in \figref{fig:human_point_cloud}. Here, we indicate some points by the red box which are not assigned the corresponding points. Note that the input point clouds are in one-to-one correspondence, thus the problem of one-to-many matching exists in baselines. Our method achieves the real one-to-one matching with high matching precision.

For a fair comparison, the proposed method is compared with those 3D mesh-based methods on re-meshed FAUST and SCAPE. Here, we use the vertices as our input point cloud. And the per-point-average geodesic distance results between the ground truth and predicted corresponding points are reported. Following \cite{donati_deepgfm_cvpr_20}, all results are multiplied by 100 for the sake of readability. 
From the \tabref{tab:fs}, our method gives accurate results on all settings, which is comparable with the state-of-the-art method, DeepGFM \cite{donati_deepgfm_cvpr_20} outperforming other baselines. Moreover, we speculate that when the deformation becomes more severe, \eg SCAPE dataset, the mesh-based method DeepGFM is more effective, where the topology information plays a more important role.

\begin{figure}[!t]
	\centerline{\includegraphics[width=7cm]{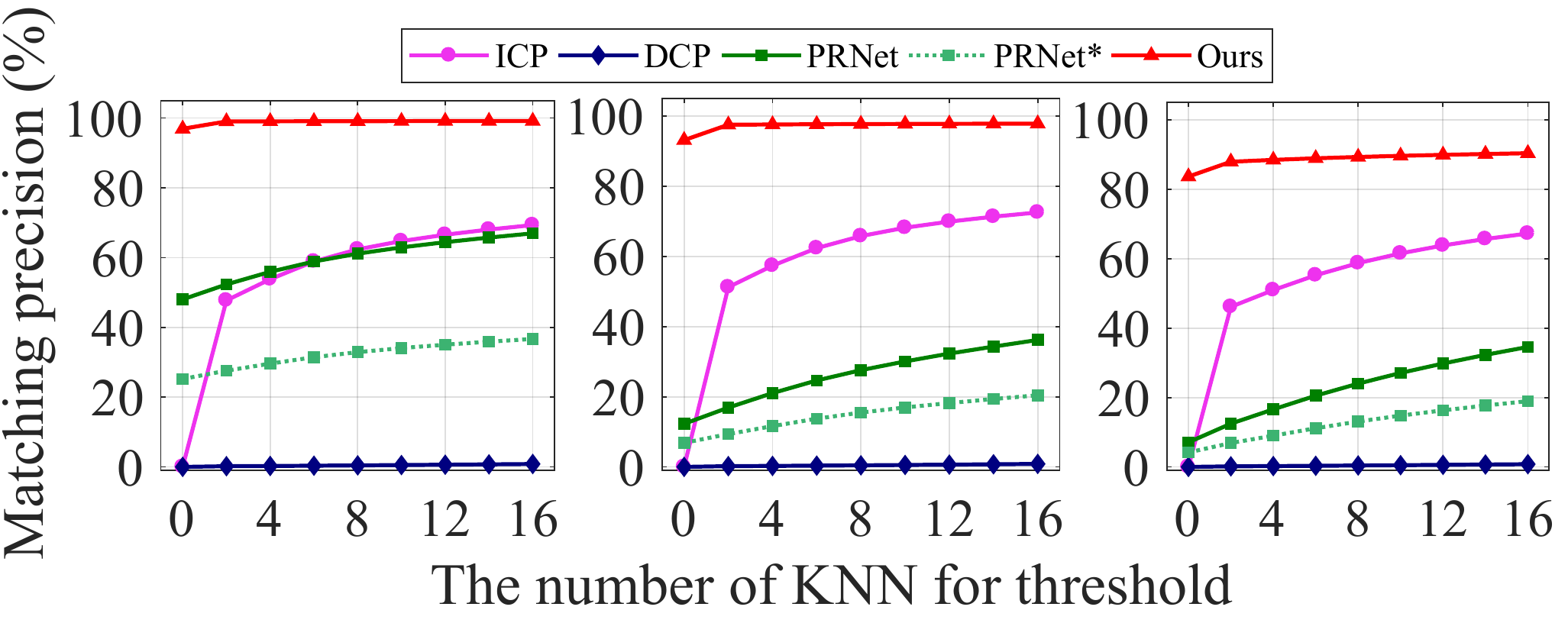}}
	\vspace{-0.2cm}
	\caption{Matching precision on rigid data. Three sub-figures indicate three settings, \ie \textit{UPC}, \textit{UC} and \textit{ND}. Obviously, our method outperforms baselines with a big margin.}
	\vspace{-0.6cm}
	\label{fig:match}
\end{figure}

\subsection{Evaluation on Rigid Data}
\label{sec:exp:rigid}
In this section, we evaluate the dense correspondence estimation on rigid point clouds. For a comprehensive comparison, we also report the rigid transformation estimation results.

Following \cite{wang_dcp_iccv_2019,yew_rpmnet_cvpr_2020}, three different experiment settings are employed here. 1) {Unseen point clouds} (\textit{UPC}): all the point clouds in the ModelNet40 are divided into training and test sets with the official setting. 2) {Unseen categories} (\textit{UC}): in ModelNet40, we select the first 20 categories for training and the rest for testing. 3) {Noisy data} (\textit{ND}): random Gaussian noise (\ie, $\mathcal{N}(0,0.01)$) is added to each point of input point clouds, while the sampled noise out of the range of $[-0.05,0.05]$ will be clipped. The dataset split strategy is the same as \textit{UPC}.

\noindent\textbf{Correspondence estimation.}
\label{sec:exp:rigid:correspondences}
We draw the matching precision in \figref{fig:match}. The proposed method outperforms other state-of-the-art methods with a big margin. These obtained results are reasonable because the DCP-v2 is a virtual point-based method that fails to predict the corresponding points and PRNet and PRNet* have serious one-to-many matching problem.

\noindent\textbf{Rigid transformation estimation evaluation.}
We also provide a comparison of transformation estimation, which is achieved by the Procrustes algorithm \cite{gower_procrustes_1975} based on the learned correspondence. 
Following \cite{wang_dcp_iccv_2019,yew_rpmnet_cvpr_2020}, we use the metrics of root mean square error (RMSE) and mean absolute error (MAE) between the ground truth and prediction of the Euler angle and translation vector, notated as RMSE(R), MAE(R), RMSE(t), and MAE(t). 
Among all the experimental settings in \tabref{tab:registration}, our method achieves the best results on all metrics.

\begin{table}[!t]
	\renewcommand\arraystretch{1.0}
	\setlength\tabcolsep{3pt}
	\caption{Evaluation of transformation estimation.}
	\vspace{-0.3cm}
	\begin{center}
		\resizebox{255pt}{!}{
			\begin{tabular}{lcccccccccccc}
				\toprule
				\multirow{2}*{\textbf{Method}}&\multicolumn{3}{c}{\textbf{RMSE(R)}}&\multicolumn{3}{c}{\textbf{MAE(R)}}&\multicolumn{3}{c}{\textbf{RMSE(t)}\ ($\times 10^{-4}$)}&\multicolumn{3}{c}{\textbf{MAE(t)}\ ($\times 10^{-4}$)} \\
				\cmidrule(r){2-4} \cmidrule(r){5-7} \cmidrule(r){8-10} \cmidrule(r){11-13}
				~&\multicolumn{1}{c}{\textit{UPC}}&\multicolumn{1}{c}{\textit{UC}}&\multicolumn{1}{c}{\textit{ND}}
				 &\multicolumn{1}{c}{\textit{UPC}}&\multicolumn{1}{c}{\textit{UC}}&\multicolumn{1}{c}{\textit{ND}}
				 &\multicolumn{1}{c}{\textit{UPC}}&\multicolumn{1}{c}{\textit{UC}}&\multicolumn{1}{c}{\textit{ND}}
				 &\multicolumn{1}{c}{\textit{UPC}}&\multicolumn{1}{c}{\textit{UC}}&\multicolumn{1}{c}{\textit{ND}} \\
				\midrule
				\textbf{ICP}       &12.282 &12.707 &11.971 &4.613 &5.075 &4.497 &477.44 &485.32 &483.20 & 22.80 &23.55 & 23.35\\
				\textbf{FGR}       &20.054 &21.323 &18.359 &7.146 &8.077 &6.367 &441.21 &457.77 &391.01 &164.20 &18.07 &144.87\\
				\textbf{PTLK}      &13.751 &15.901 &15.692 &3.893 &4.032 &3.992 &199.00 &261.15 &239.58 & 44.52 &62.13 & 56.37\\
				\textbf{DCP-v2}    & 1.094 & 3.256 & 8.417 &0.752 &2.102 &5.685 &17.17  & 63.17 &318.37 & 11.73 &46.29 &233.70\\
				\textbf{PRNet}     & 1.722 & 3.060 & 3.218 &0.665 &1.326 &1.446 &63.72  &100.95 &111.78 & 46.52 &75.89 & 83.78\\
				\textbf{PRNet*}    & 2.090 & 3.720 & 3.292 &0.894 &1.543 &1.449 &109.79 &131.33 &107.68 & 81.49 &99.61 & 81.51\\
				\midrule
				\textbf{Ours}      &\textbf{0.864}&\textbf{1.962}&\textbf{3.006}&\textbf{0.114}&\textbf{0.338}&\textbf{0.854} 
				                       &\textbf{1.45}&\textbf{4.09}&\textbf{5.72}&\textbf{0.12}&\textbf{0.43}&\textbf{1.03}\\
				\bottomrule
			\end{tabular}}
		    \label{tab:registration}
	\end{center}
	\vspace{-0.5cm}
\end{table}

\subsection{Ablation Studies}
\label{sec::ablation}
\begin{figure}[!t]
	\centering
	\vspace{-0.1cm}
	\includegraphics[width=7cm]{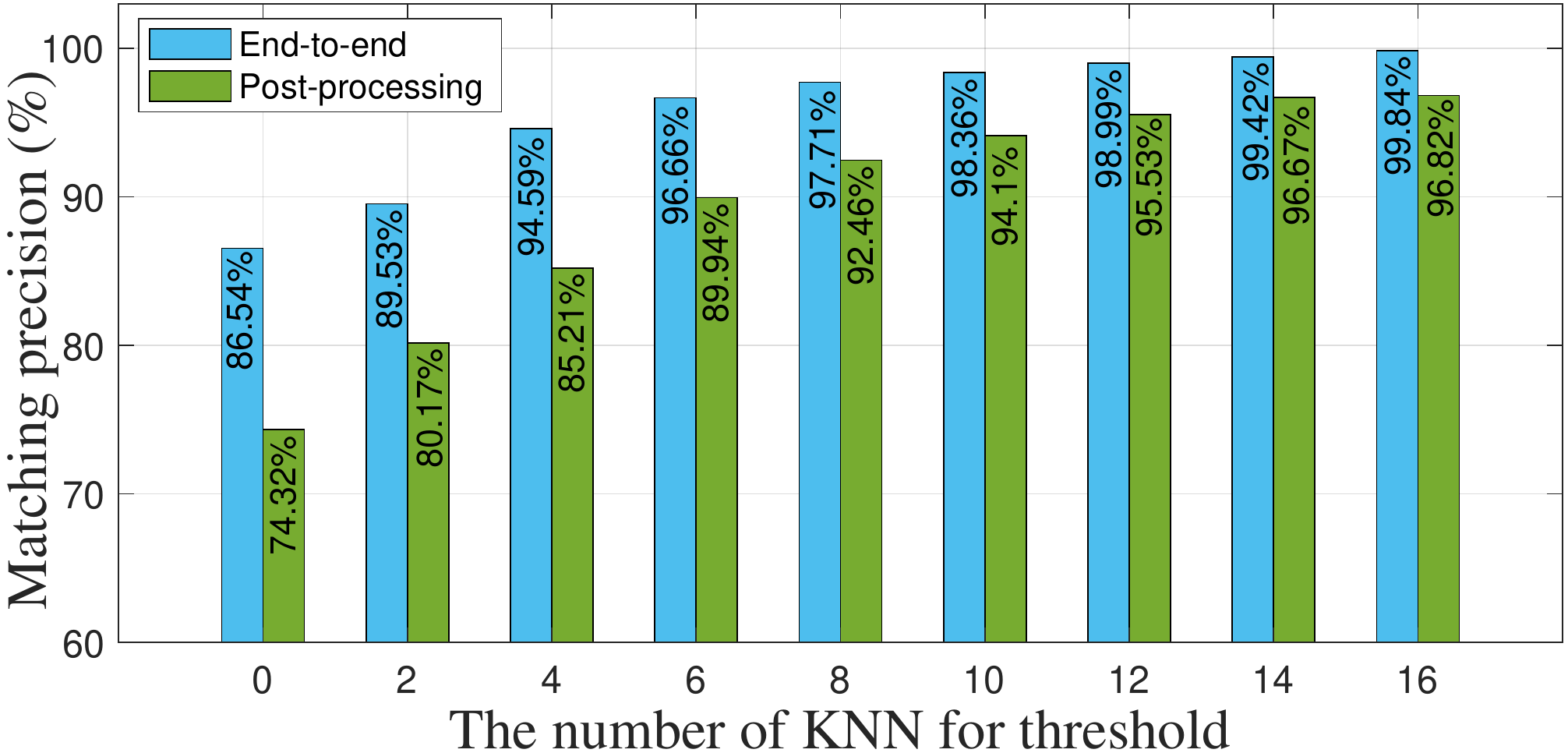}
	\vspace{-0.15cm}
	\caption{Comparison of post-processing, end-to-end learning. %
	}
	\vspace{-0.50cm}
	\label{fig:abla}
\end{figure}

\noindent\textbf{End-to-end training vs. Post-processing.}
We solve the zero-one assignment problem by end-to-end learning. Another more direct idea is to solve this non-convex problem by post-processing, \ie learning the soft matching matrix and solving the permutation matrix during the test. From the comparison provided by \figref{fig:abla}, which is conducted on Surreal\textbackslash SHREC, the end-to-end learning benefits the final results significantly because more accurate loss is calculated during the training. 

\section{Conclusion} \label{sec:conclusion}

In this work, we advocate achieving one-to-one matching by learning the permutation matching matrix to tackle the dense correspondence estimation problem for 3D point clouds.
Moreover, to guarantee the end-to-end learning and the accuracy of the calculated loss, an ingenious matching module has been designed in our method.
Extensive experiments on both non-rigid and rigid 3D point cloud data have validated that ours achieved state-of-the-art performance. 
However, our method has limitations. Specifically, our method essentially determines the correspondences according to the similarity of the points. Thus, ours cannot deal with objects with strong symmetry, which is a stubborn problem in the community.
In the future, we plan to extend our method to other dense correspondence estimation tasks, \eg 2D-2D, 2D-3D.

\bibliographystyle{IEEEtran}     
\bibliography{Learning_permutation_matrix_Reference.bib}
\end{document}